%% file: main.tex
\documentclass[11pt]{article}
\input{_preamble}

\pdfoutput=1
\usepackage{mathptmx}
\usepackage{soul}

\newcommand\titl{Are Chatbots Reliable Text Annotators? Sometimes}
\title{\titl%
  \thanks{
  Acknowledgements: We thank M. W. Kolster, A. L. Kristensen, and T. Buntzen for annotation, P. B. Vahlstrup for data collection and Independent Research Fund Denmark for funding (Grant nr. 0133-00187B). All computation was performed on the UCloud interactive HPC system, managed by the SDU eScience Center.
  }
}

\author{
Ross Deans Kristensen-McLachlan\footnote{Corresponding Author. Center for Humanities Computing, Aarhus University, Denmark, rdkm@cc.au.dk}\\\vspace{-1em}
Miceal Canavan\footnote{Department of Political Science, Aarhus University, Denmark.}\\
Márton Kardos\footnote{Center for Humanities Computing, Aarhus University, Denmark.}\\
Mia Jacobsen\footnote{Center for Humanities Computing, Aarhus University, Denmark.}\\
Lene Aarøe\footnote{Department of Political Science, Aarhus University, Denmark.}
}

\date{\normalsize \today}

\begin{document}
\pagenumbering{gobble}

\maketitle
\onehalfspacing

\begin{abstract}

Recent research highlights the significant potential of ChatGPT for text annotation in social science research. However, ChatGPT is a closed-source product which has major drawbacks with regards to transparency, reproducibility, cost, and data protection. Recent advances in open-source (OS) large language models (LLMs) offer an alternative without these drawbacks. Thus, it is important to evaluate the performance of OS LLMs relative to ChatGPT and standard approaches to supervised machine learning classification. We conduct a systematic comparative evaluation of the performance of a range of OS LLMs alongside ChatGPT, using both zero- and few-shot learning as well as generic and custom prompts, with results compared to supervised classification models. Using a new dataset of tweets from US news media, and focusing on simple binary text annotation tasks, we find significant variation in the performance of ChatGPT and OS models across the tasks, and that the supervised classifier using DistilBERT generally outperforms both. Given the unreliable performance of ChatGPT and the significant challenges it poses to Open Science we advise caution when using ChatGPT for substantive text annotation tasks. 

\end{abstract}
\newpage    

\doublespacing
\par The rapid development of large language models (LLMs) such as ChatGPT has generated substantial scientific interest. Proponents suggest they show promise in solving research challenges, such as large scale text annotation \parencite{gilardi23,heseltine24, rathje23}. However, this still nascent technology has significant drawbacks, especially with regards to Open Science. It is therefore important to evaluate their performance to: 1) advance understanding of their accuracy and reliability of measurement; 2) clarify to what extent researchers face a trade-off between classification accuracy and Open Science principles; and 3) stimulate discussion about whether such a trade-off is desirable or acceptable. In this paper, we systematically evaluate the performance of ChatGPT against alternative open source (OS) models and supervised classifiers developed on a new set of data and tasks.

\par Early experimentation has highlighted the potential of ChatGPT for text annotation \parencite{gilardi23,heseltine24, rathje23}, but the empirical evidence is still limited and other studies are more circumspect about the accuracy and reliability \parencite{yu23,ollion23}. A preliminary review concluded “evidence about their effectiveness remains partial" \parencite[p. 1]{ollion23}. Furthermore, ChatGPT is a closed-source, proprietary model which raises significant open science concerns with respect to transparency and reproducibility, alongside cost barriers and data protection issues \parencite{wu23,ollion23}. 

\par These limitations can be overcome with both supervised classifiers and OS LLMs. These approaches make the underlying model openly available, can be used free of charge, and run on local hardware to mitigate data privacy and protection issues. Furthermore, recent research suggests that OS LLMs may “represent a competitive alternative for text annotation tasks, exhibiting performance metrics that generally exceed those of MTurk and rival those of ChatGPT” \parencite[p. 6]{alizadeh23}. Other research, however, finds that the current generation of LLMs, including both OS models and ChatGPT can be unreliable and unpredictable \parencite{yu23}. While some technical solutions have been proposed as a remedy these problems \parencite{lee-etal-2023-making}, there has been limited side-by-side comparison of these three potential solutions (i.e., open-source LLMs, closed-source LLMs, and supervised classifiers) on novel data and tasks. Such a comparison is important because it allows for a direct comparison of classification performance. 

\par We systematically evaluate the performance of contemporary OS LLMs and multiple ChatGPT models, with two additional supervised machine learning classifiers trained by the authors on human-labeled data. We evaluate model performance on two different binary classification tasks on a new dataset of tweets from US news media (n=1,000 tweets for each task), experimenting with few-shot and zero-shot learning, and using generic and custom prompts. To increase ecological validity, we focus on binary tasks because they have broad applicability across the social sciences and related fields (see Materials and Methods for model architectures, data, and tasks). We choose not to fine-tune the LLMs since this can be technically challenging, requiring resources which are unavailable for most researchers. Our aim is to achieve a common ecological validity, focusing on how this technology is most likely to be used "in the wild".

\par Overall, our findings suggest that researchers do not face a trade-off between classification accuracy and Open Science principles, since both open- and closed-source LLMs tend to underperform relative to certain supervised machine learning approaches, irrespective of model or prompting procedure. ChatGPT generally outperforms OS models, but we find significant inconsistency and unreliability across different tasks and models. Given the closed-source nature of ChatGPT, it is challenging to pinpoint the cause of this inconsistency. We conclude with recommendations for researchers on using LLMs for text annotation. 

\vspace{5mm} 

\section*{Results}

\par In our evaluations, we report both accuracy and F1. Accuracy, meaning the percentage correct agreement relative to ground truth, is widely used and intuitive to interpret but is typically inflated when classes are unbalanced. Conversely, F1 is a more robust performance evaluation metric for unbalanced data, representing the harmonic mean of precision and recall. Full results of all models and classification reports can be found in the associated Harvard Dataverse repository.

\par In Figure 1 below, panel A shows the accuracy and panel B shows the F1-score of all models across all permutations. In each panel, the two columns represent the individual binary classification tasks, and the rows represent models with custom and generic prompts, respectively. The categories are the presence of political/non-political and exemplar/non-exemplar content in the US news media tweets. 

\vspace{5mm} 

\par \begin{center} \textit{Figure 1 about here.}\\
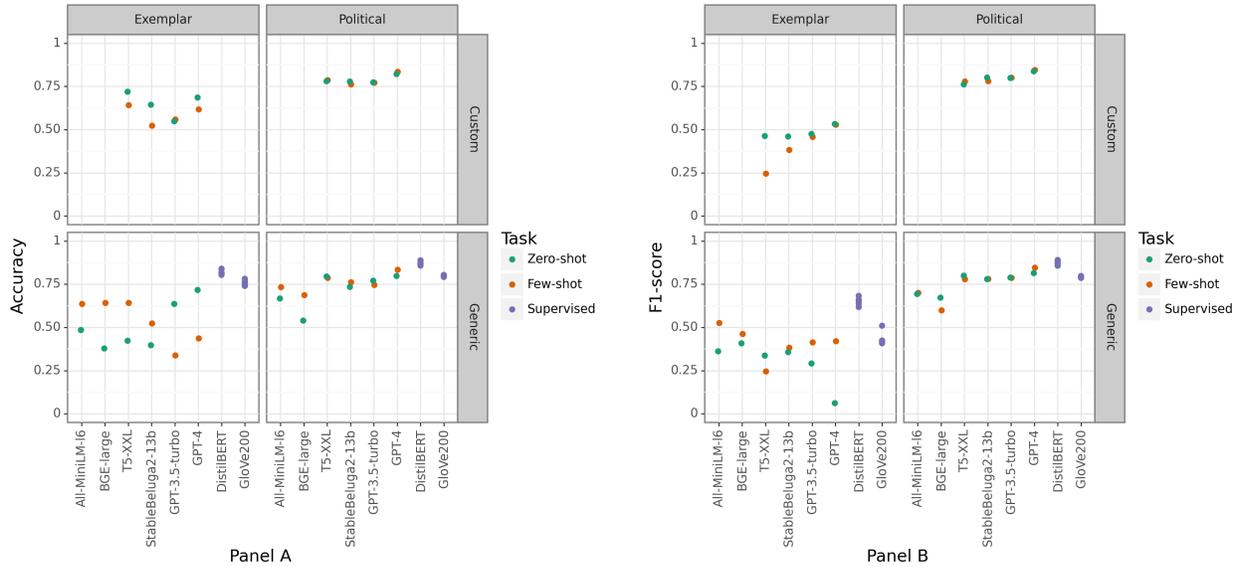
\textbf{Figure 1:} \textit{Accuracy (i.e. percentage of correct predictions) and F1-scores by model type, prompt type, and zero- or few-shot.}
\end{center}

\vspace{5mm} 

\par The supervised classifier using DistilBERT has the highest performance of all the models on both tasks, matched only by GPT-4 for predicting political content. No other model comes close to DistilBERT's performance for the exemplar task. FLAN-T5-XXL performs well in zero- and few-shot contexts on political labels with both custom and generic prompting. However, the same model performs worst for predicting exemplar label across all permutations.

\par The supervised model trained with static GloVe embeddings performed notably worse on predicting political label. Moreover, performance is generally worse on the exemplar task for all models regardless of prompt structure or zero- versus few-shots learning with extreme variation in performance on this task. Perhaps most notably, GPT-4 performs significantly \textit{worse} than any other model in the context of zero-shot labeling on exemplar data.  

\par Our results show that, for both tasks, only the supervised DistilBERT model has consistently and meaningfully high enough performance above chance to be considered useful. Moreover, custom prompts seem generally to generate marginally higher performance than generic prompts for both ChatGPT and OS models for the exemplar task. However, most importantly LLM performance on both tasks varies in unpredictable and inconsistent ways depending on the prompt and zero- or few-shot approach.

\par Figure 2 expands on these results, reporting the recall and precision scores for the positive classes (i.e. `political' and `exemplar'). These results further support that the supervised classifiers mostly have a higher performance. Furthermore, the performance of the ChatGPT and OS models cluster around the same good-to-high performance levels for the political class with relatively few percentage-point deviance. However, for the exemplar class all LMMs have equally low precision, with greater variation in recall. Most strikingly, GPT-4 demonstrates recall as low as 0.1 when classifying exemplar tweets (zero-shot, generic prompt). 

\par The intercoder reliability scores for the multicoded data indicates strong intercoder reliability among the human coders - $\alpha$=0.86 (political) and 0.84 (exemplar) . Since humans did not reach perfect agreement, it may be unrealistic to expect tools for automated classification to reach perfect accuracy. Nevertheless, the variation in Figure 1 suggests that the performance of LLMs on particular tasks is not easily predictable from human performance.

\vspace{5mm} 

\par \begin{center} \textit{Figure 2 about here.}\\
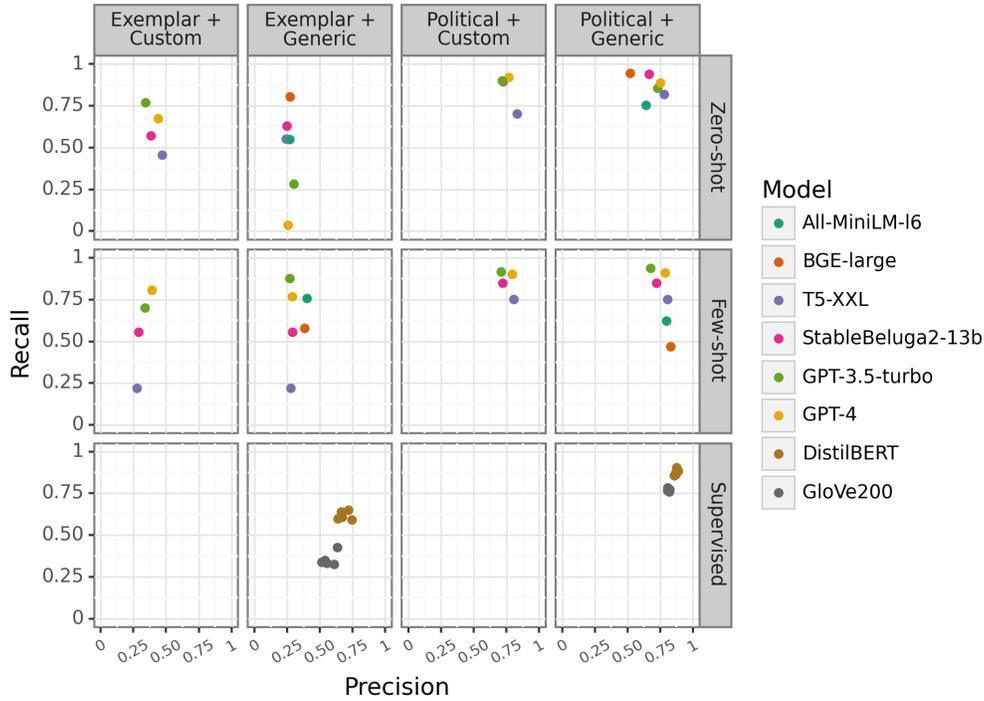
\textbf{Figure 2:} \textit{Precision and recall score of positive class by model type and prompt.}
\end{center}

\vspace{5mm} 

These results are significant for a number of reasons. First, although the performance of the LLMs is impressive in places, the supervised machine learning approach -- especially DistilBERT -- generally performs best. Second, while GPT-4 in particular marginally outperforms OS LLMs, in other instances the OS models are within a few percentage points of this model or even outperform it. Additional prompt engineering may improve performance, and testing this iteratively is significantly simpler and cheaper when using a smaller models like Llama3.1-8b which can be run locally, rather than paying on a token-by-token basis for API access to ChatGPT.

\vspace{5mm} 

\section*{Discussion and recommendations}

\par The performance of ChatGPT and other OS LLMs varies significantly and often unpredictably on our binary classification task. The different OpenAI models in particular demonstrate significant variation and the source of this is opaque, since OpenAI's closed-source policies make testing and evaluation impossible. Closed models slightly outperform OS LLMs on certain tasks, but both are consistently matched or outperformed by the supervised classifier using DistilBERT. These results are more cautious about the potential of ChatGPT than earlier studies \parencite{alizadeh23,gilardi23, heseltine24} and align with more cautionary conclusions \parencite{yu23,ollion23}.

\par Furthermore, OpenAI (and similar) continuously adjust their models, meaning that research and annotation conducted using ChatGPT today may not be directly replicable or even reproducible in six months. Our findings indicate that these challenges to Open Science come with limited performance gains compared to OS alternatives and typically none compared to the supervised classifier. On this basis, we advise caution in using ChatGPT for text annotation in social science research at this time and conclude that the most consistent way for researchers to perform reliable, transparent, and efficient text annotation at scale is through human-labeling of data and training supervised classifiers. 

A limitation of our study is that our results are based on a single dataset in English, meaning generalizability may be limited. Recent research finds that ChatGPT performs similarly or (slightly worse) in other languages \parencite{rathje23} and other types of text data \parencite{gilardi23}. Our analysis also does not include fine-tuning of LLMs, which may improve performance. We find that custom prompting and few-shot examples generally improves performance but the effects are inconsistent and further research is needed.

While Generative AI continues to capture both popular imagination and the attention of researchers, our results demonstrate that LLMs are currently too brittle and potentially unreliable for scientific research. Moreover, closed-source versions are opaque and prohibitively expensive when working with large datasets. We argue that research time and finances are instead better invested in the development and application of encoder-only models such as DistilBERT which can be integrated into standard supervised learning pipelines. While perhaps less immediately glamorous, these models continue to match or outperform decoder-only models for a fraction of the cost and with the added bonus of greater transparency and reproducibility.

\vspace{5mm} 

\subsection*{Data availability statement} \par The data that support the findings of this study are openly available in Harvard Dataverse (\url{https://doi.org/10.7910/DVN/TM7ZKD}). Data are shared in accordance with Twitter/X’s Developer Agreement, meaning that only post IDs can be shared, not the tweets themselves. All Python code to reproduce our results can be found on available on Github and Zenodo at (\url{https://zenodo.org/records/14916594}).

\subsection*{Materials and methods}  
\par \textit{Data collection and annotation:} We use a dataset of tweets from US news media outlets collected from 2021 to 2023 by the authors for a new research project. These tweets were annotated for whether or not they contained political content, and whether or not they contained an exemplar. All tweets were annotated by three research assistants. Any discrepancies were subsequently resolved by an expert coder (an author of this paper). On the multi-coded data, the intercoder reliability between three research assistants was high (0.86 for political, 0.84 for exemplar), therefore, there was a clear shared understanding even before discrepancies were resolved. For each task, we have 1,000 multi-coded tweets. This constitutes the ground truth data that was used to train the supervised classifier and to assess performance. From each of these 1,000 multi-coded datasets, a fixed  sample of 200 tweets was drawn for testing the LLM-based models. 

\par \textit{Supervised machine learning classifiers:} Human-coded tweets were first transformed into dense numerical representations known as word embeddings. We train models using both static (GloVe) and contextual word embeddings (DistilBERT) \parencite{devlin18,sanh19}. These classifiers were trained using stratified K-fold cross validation (k=5). 

\par \textit{LLMs:} We test a range of possible zero-and few-shot learning models with basic prompting. The models were selected based on 1) past classification performance and 2) widespread usage at time of writing and 3) accessibility (see details on prompt instructions in the Supplementary Material). Firstly, we use two sentence-transformers \parencite{reimers19} which had demonstrated high levels of performance across a number of NLP tasks at the time of the initiation of the research: \textit{all-minilm-L6}; and \textit{bge-large-en} \parencite{muennighoff22}. For OS LLMs, we use \textit{FLAN-T5-XXL} \parencite{chung22}, a Text-to-Text model released by Google based on their earlier T5 architecture which has shown promising classification performance \parencite{alizadeh23}. We also included autoregressive, decoder-only transformer models Llama-3.1-8B created by Meta and StableBeluga2-13b, a fine-tune of an earlier Llama2 model created by StabilityAI. Finally, we test two LLMs underlying ChatGPT, namely GPT-3.5-turbo and GPT-4, accessed via the OpenAI API.

\clearpage
\singlespacing
\printbibliography

\clearpage
\newpage

\section*{S1. Codebook and Human Annotation}

\subsection*{US News Media Tweet Dataset} Using the Twitter API we scraped tweets from approximately 360 American news media organizations from January 9, 2021, to January 4, 2023, for a new research project on political news communication on Twitter. In this report we analyze a random sample of 2,000 multi-coded tweets. The sample size of 2,000 tweets (1,000 for each code) was determined to ensure a reasonable number of tweets for training supervised machine learning classifiers with our cross validation strategy.

\subsection*{Annotation} Using a codebook, three student research assistants were trained to label tweets for the political and exemplar tasks (see codebook guidelines below). These tasks were chosen specifically since one is a broadly understood concept with a shared meaning, whilst the other is a more multivalent concept used infrequently in everyday language. All tweets were labeled by three research assistants where discrepancies were resolved by an expert coder (a co-author of this article). In the data annotated by three research assistants, there was high intercoder reliability between annotators - 0.86 for political and 0.84 for exemplar before the remaining discrepancies were resolved. These 2,000 tweets (1,000 for each code) constitute our gold standard for the political and exemplar task. This approach is in line with leading research on this topic - it minimizes false negatives/positives where the human label is wrong, and therefore gives a more accurate reflection of the performance of LLMs and supervised models.\footnote{See for example Umansky, N., Kubli, M., Donnay, K., Gilardi, F., Hangartner, D., Kotarcic, A., ... and Grech, P. Enhancing Hate Speech Detection with Fine-Tuned Large Language Models Requires High-Quality Data. DOI:10.31219/osf.io/7kbqt}

\par For the exemplar task, our data comprises 432 labeled as exemplar and a further 568 labeled as non-exemplar (approximately 1:1 ratio). For the political task, there are 247 tweets labeled as political and 753 as containing non-political (approximately 1:3 ratio).  

\subsection*{Codebook Guidelines} We annotated for two tasks, political and exemplar. Below we report the definition for the political and exemplar task provided in the codebook. The codebook also included examples of tweets from each category and operational annotation instructions (code definition). The limited number of tweets concerning international news was labeled as "non-exemplar" and "non-political" as the broader research project for which the codebook was developed focuses on US domestic politics. The full codebook for the two tasks is available on the associated Dataverse for this report.   

\textbf{Political Task:} Given we are focused on social media, the starting point for identifying political content builds on Twitter’s definition of political content which it uses to regulate political content: ‘\textit{Political content is defined as content that references a candidate, political party, elected or appointed government official, election, referendum, ballot measure, legislation, regulation, directive, or judicial outcome.}’ 

\par Another definition which can be used as a guide was developed by Petersen (2018, p.5)\footnote{Petersen, H.H. (2018). Data Report: Facebook Posts of Danish MPs, \url{https://ps.au.dk/fileadmin/ingen\_mappe\_valgt/Facebook\_data\_report\_May\_2018.pdf}} for analysing Facebook posts by politicians: ‘\textit{Political content includes ‘all statements with a political substance, but excludes messages solely related to non-political matters such as birthdays or personal experiences}’. The message can express a political attitude directly or indirectly. For instance, a message saying “Thank you Emsworth Primary School for a nice visit” does not count as a political message, but if the message is followed with a political statement eg. “let’s unite to secure good education for all” it should be coded as political. Further, political messages can comment on political processes and practices such as: “The government does not do what it takes to make the parties cooperate”. Hence, a political message will generally state a policy position, raise awareness of a political cause, encourage people to take action to promote a political goal or reflect upon a political decision-making process.

\par \textbf{Exemplar Task:} Journalists often utilize examples of people to illuminate the story they are reporting on. Conceptually, these people are referred to as ‘exemplars’. For the purposes of this research, we adopt a broad definition of exemplars, as people who are mentioned in relation to the story being reported. They can be people directly affected and involved in the issue, or they may simply be commenting on the issue. This broader definition which includes unaffected and affected exemplars is illuminated in further detail by Zerback \& Peter (2018).\footnote{Zerback, T., \& Peter, C. (2018). Exemplar effects on public opinion perception and attitudes: The moderating role of exemplar involvement. Human Communication Research, 44(2), 176-196.}

This definition expands on the classic understanding in the communication literature which required that the person portrayed be an active part of the story or directly affected. As a result, this expanded definition includes so-called ‘vox populi’ where people are simply giving their opinion on an issue or policy being reported.

\vspace{5mm} 

\section*{S2. Model selection}
As mentioned in the main text we test a range of possible zero- and few-shot learning models with basic prompting to evaluate the performance of LLMs. As mentioned in the main text, the models were selected based on 1) past classification performance and 2) widespread usage, and 3) accessibility. The OS LLMs included represent models released since late 2023 which have shown high-performance and wide user adoption. That the models tested include both examples from the beginning of this period of time and more recent OS LLMs increases the external validity.

As noted in the main text we do not fine-tune the LLMs on the task, since this is a technically challenging task requiring resources and which may therefore not be available to many scholars. Nevertheless, the degree of
technical complexity of fine-tuning depends on the type of model, and recent research finds that fine-tuning can improve LLM performance and this is an important point for future research.  \footnote{See for example Umansky, N., Kubli, M., Donnay, K., Gilardi, F., Hangartner, D., Kotarcic, A., ... and Grech, P. Enhancing Hate Speech Detection with Fine-Tuned Large Language Models Requires High-Quality Data. DOI:10.31219/osf.io/7kbqt and Alizadeh, M., Kubli, M., Samei, Z., Dehghani, S., Bermeo, J. D., Korobeynikova, M., \& Gilardi, F. (2024). Open-source large language models outperform crowd workers and approach ChatGPT in text-annotation tasks. arXiv preprint arXiv:2307.02179v2.}

\vspace{5mm} 

\section*{S3. Few shot learning examples}
In the few-shot approach, we provide the LLMs with some examples of tweets which are considered US politics and exemplars, and those which are not. These examples were randomly selected from tweets which all research assistant expert coders agreed represented the category in the training data. Random selection gives us breadth of coverage of different types of tweets in each category and using only tweets where all three research assistants agree gives us validity they capture the concept. The examples we provided were:

\textbf{Political:}
\begin{itemize}
    \item `In my opinion, to be faithful is to tell both sides of the story.' GOP Sen. Tim Scott discusses a part of his response to Biden's address that almost didn't happen and race relations in America. https://t.co/WbjnFJEibq
    \item A New Day: Why Neither Politicians Nor Clergy Can Hide From Abortion Any Longer https://t.co/jxFQKfTldr
    \item Finland and Sweden have mostly stopped prescribing blockers to under-18s in favour of talking therapy, because the evidence base for them is thin. Joe Biden’s order asks federal departments to expand access to “gender-affirming care” https://t.co/cmylNEJ0Gj 
    \item'We're broken.' In the suburbs north of Los Angeles, voters feel fed up and afraid https://t.co/Gzxc6zlhnS
    \item House Democrat Chris Pappas is in a close reelection race in New Hampshire. Good news: Sabato just upgraded the race from “Toss up” to “Lean Democrat.” But it’ll still be close. Donate right now and help put @ChrisPappasNH over the top: https://t.co/QjHgEpVXl4
\end{itemize}

\textbf{Non-political (including international):}
\begin{itemize}
    \item Rialto police apologize to teenage girl's family after violent arrest is caught on video https://t.co/xdZhIaKthC
    \item Following months during which Kanye West had been publicly addressing his divorce and co-parenting, he has reportedly told Kim Kardashian that he is `going away to get help.'  https://t.co/VbubgWmkdS
    \item Expect the return of holiday favorites, including the beloved classic `Santa Claus Is Comin' to Town' and a special airing of `Home Alone' on Christmas Eve.  https://t.co/w6kiFc8JDM
    \item Haaretz reports that according to some witnesses, the crowd celebrating Itamar Ben-Gvir’s election to the Israeli Knesset “chanted ‘death to Arabs’ alongside the more prevalent calls for ‘death to terrorists.’” https://t.co/9N4rwBybU7
    \item \#ANALYSIS: 'Is your leadership safe$?$': The question that shows how much Morrisons position has fallen https://t.co/ApJJLCc1JC
\end{itemize}

\textbf{Exemplar:}
\begin{itemize}
\item `The twists and turns in Spears’ story over recent years have fundamentally altered the dream of becoming a pop star, even as the appeal of finding one artist who can make a song that changes the world for five minutes remains' writes @maura https://t.co/os1SH3Drsg
\item RT @TedAbram1: YES!!! House Minority Leader Kevin McCarthy said Republicans are launching an investigation to find out why American tax dol\footnote{Due to the scraping procedure for retweets, the final characters of the original tweet can be cut, we still code these tweets where there is enough information to indicate the code}
\item Two men rescued after 29 days lost at sea, survived on oranges and rainwater https://t.co/f5GGziG3vr https://t.co/RxlCNvJ4Xe
\item MSNBC analyst takes up arms in Ukraine because he's 'through talking about it' https://t.co/5DvDO2780C
\item RT @emilytgreen: “They left him as if he were—I won’t say an animal, because animals are sentient beings,” the family’s lawyer said. “As if…
\end{itemize}

\textbf{Nonexemplar (including international):}
\begin{itemize}
\item Operation Holiday Cheer sends Christmas trees to troops | https://t.co/EZaojUcXuX
\item Border control advocates want Americans to snitch on ICE’s ‘secret night flights’  https://t.co/mMNXVmE7vg https://t.co/iJtgNlV1wa
\item RT @MikeyNoWay: \#Spinout on WB Hwy 24 near Wilder Rd. leaves right lane blocked by crews for overturned vehicle. A second crash tied up tra…
\item The Healthy School Meals for All program is an investment in public education and in our children who need healthy food to learn. https://t.co/RL75NzNjx3
\item https://t.co/7clmW8jHe3\footnote{Some of the tweets scraped in the dataset are URLs only. These were coded as not having an exemplars or political content present}
\end{itemize}

\vspace{5mm} 

\section*{S4. Stormtrooper package}
To perform comparable tests (both the original analysis and replication), we utilized the Python package stormtrooper (Kardos, 2023) \footnote{Kardos, M. (2023). stormtrooper: scikit-learn compatible zero and few shot learning in Python, \url{https://github.com/centre-for-humanities-computing/stormtrooper}} which is designed to perform zero- and few-shot learning with transformer-based models. The package has a series of default prompts for this task, which can be found in the relevant modules in the open-source GitHub repo for the package. The full set of prompt instructions for this report can be found both in the associated Dataverse and on the Github repository for the stormtrooper package. \footnote{Complete documentation for the stormtrooper package can also be found at the following URL: \url{https://centre-for-humanities-computing.github.io/stormtrooper/}}

For example, the default zero-shot prompt for generative decoder-only models such as StableBeluga-13B the generic prompt template is as follows:

\vskip 0.5cm
\noindent \textbf{\#\#\# System:}
\textit{\newline You are a classification model that is really good at following instructions and produces brief answers that users can use as data right away. \\ Please follow the user's instructions as precisely as you can.}
\textbf{\#\#\# User:}
\textit{\newline Your task will be to classify a text document into one of the following classes: \\
$\{CLASSES\}$.\\ 
Please respond with a single label that you think fits the document best. \\ Classify the following piece of text:}
\newline $\{INPUT\_TEXT\}$\\
\textbf{\#\#\# Assistant:}
\vskip 0.5cm

For a text-to-text model such as FLAN-T5-XXL, the generic prompt is much simpler:
\vskip 0.5cm
\noindent \textit{I will give you a piece of text. Please classify it as one of these classes: {classes}. \\ Please only respond with the class label in the same format as provided.}\\
$\{INPUT\_TEXT\}$
\vskip 0.5cm

\section*{S5. Prompting} The user of stormtrooper is free to design their own custom prompts which are more specifically targeted towards the task at hand through a process of prompt engineering. We hence also experimented with custom prompts which included definitions of the categories we with to annotate. A full overview of prompts is reported on Github. Examples of these custom prompts for StableBeluga-13B are as follows:

\bigskip
\textbf{Zero-shot custom prompt:}

\vskip 0.5cm
\noindent \textbf{\#\#\# System:}\\
\textit{You are a classification model that is really good at following instructions and produces brief answers that users can use as data right away.\\
Please follow the user's instructions as precisely as you can.}\\
\textbf{\#\#\# User:}\\
\textit{Your task will be to classify a text document into one of the following classes: \\
$\{CLASSES\}$.\\
Political content is defined as content that references a candidate, political party, elected or appointed government official, election, referendum, ballot measure, legislation, regulation, directive, or judicial outcome.\\
Political content includes all statements with a political substance, but excludes messages solely related to non-political matters such as birthdays or personal experiences.\\
Please respond with a single label that you think fits the document best.\\
Classify the following piece of text:
\newline $\{INPUT\_TEXT\}$\\}
\textbf{\#\#\# Assistant:}
\vskip 0.5cm

\textbf{Few-shot custom prompt:}
\vskip 0.5cm
\noindent \textbf{\#\#\# System:}\\
\textit{You are a classification model that is really good at following instructions and produces brief answers that users can use as data right away.\\
Please follow the user's instructions as precisely as you can.}\\
\textbf{\#\#\# User:}\\
\textit{Your task will be to classify a text document into one of the following classes:\\
$\{CLASSES\}$.\\
Political content is defined as content that references a candidate, political party, elected or appointed government official, 
election, referendum, ballot measure, legislation, regulation, directive, or judicial outcome. \\
Political content includes all statements with a political substance, but excludes messages solely related to non-political matters such as birthdays or personal experiences.\\
Here are some examples of texts labeled by experts.
$\{EXAMPLES\}$\\
Please respond with a single label that you think fits the document best.\\
Classify the following piece of text:}\\
$\{INPUT\_TEXT\}$\\
\textbf{\#\#\# Assistant:}
\vskip 0.5cm

As before, the text-to-text models such as FLAN-T5-XXL use a simpler prompt set-up, as exemplified by the following custom few-shot prompt:
\vskip 0.5cm
\textbf{T5 few-shot custom prompt:}\\
\textit{I will give you a piece of text. \\
Please classify it into one of the following classes: $\{CLASSES\}$.\\
Political content is defined as content that references a candidate, political party, elected or appointed government official, election, referendum, ballot measure, legislation, regulation, directive, or judicial outcome. \\
Political content includes all statements with a political substance, but excludes messages solely related to non-political matters such as birthdays or personal experiences.\\
Here are some examples of texts labeled by experts.\\
$\{EXAMPLES\}$\\
Please only respond with the class label in the same format as provided.
Label this piece of text:\\}
$\{INPUT\_TEXT\}$

\end{document}

%% file: _preamble.tex
\usepackage{geometry}
\usepackage{amsfonts}
\usepackage{amsmath}
\usepackage{amssymb}
\usepackage{bigints}
\usepackage[bottom]{footmisc}
\usepackage{pifont}
\usepackage{dcolumn, multirow}
\usepackage{setspace}
\usepackage{verbatim}
\usepackage{rotating}
\usepackage{paralist}
\usepackage{latexsym}
\usepackage{amsthm}
\usepackage{fullpage}
\usepackage[compact]{titlesec}
\usepackage{titletoc}
\PassOptionsToPackage{hyphens}{url}
\usepackage[usenames,dvipsnames,svgnames,table]{xcolor}
\usepackage{arydshln}
\usepackage{pgfplots}
\usepackage{bbold}
\usepackage{fancyhdr}
\usepackage{color}
\usepackage{graphicx}
\usepackage{type1cm}
\usepackage{eso-pic}
\usepackage{marginnote}
\usepackage{setspace}
\usepackage{parskip}
\usepackage{capt-of}
\usepackage{tikz}
\usepackage{wrapfig}
\usepackage{authblk,etoolbox}
\usepackage{pdflscape}
\usepackage{subcaption}
\usepackage{bigdelim}
\usetikzlibrary{arrows}
\usepackage{adjustbox}
\usepackage{tabularx}
\usepackage[utf8]{inputenc}

\usepackage{hyperref}
\hypersetup{colorlinks=true,citecolor=Teal,urlcolor=Teal,linkcolor=black}

\usepackage[toc,page]{appendix}

\makeatletter
\patchcmd{\@maketitle}{center}{flushleft}{}{}
\patchcmd{\@maketitle}{center}{flushleft}{}{}
\patchcmd{\@maketitle}{\LARGE}{\Large}{}{}

\def\maketitle{{%
  
  \AB@maketitle}}
\makeatother

\renewcommand\Affilfont{\normalfont\normalsize}

\usepackage[font=normalsize,labelfont=bf,margin=.3in,tableposition=top]{caption}

\usepackage[all]{nowidow}

\titleformat{\part}{\large\bfseries}{\thesection}{0em}{}
\titleformat{\section}{\large\bfseries}{\thesection.}{1em}{}
\titleformat{\subsection}{\normalsize\bfseries}{\thesubsection}{1em}{}

\usepackage{accents}

\usepackage[colorinlistoftodos, textsize=tiny, textwidth=.75in, backgroundcolor=white, linecolor=red, bordercolor=red]{todonotes}

\usepackage{bbding}
\usepackage{amssymb}
\usepackage{fancyhdr}
\usepackage{graphicx}
\usepackage{xfrac}
\usepackage{textcomp}

\usepackage{colortbl}
\definecolor{lightgray}{gray}{0.95}

\usepackage{longtable}
\usepackage{array} 
\usepackage{booktabs}        
\usepackage{enumitem}
\usepackage{setspace}

\usepackage[backend=biber,authordate, maxcitenames=2, url=false]{biblatex-chicago}

\DeclareFieldFormat{titlecase}{\MakeSentenceCase*{#1}}

\DeclareFieldFormat{postnote}{#1}
\DeclareFieldFormat{multipostnote}{#1}

\usepackage{pifont}

\pgfplotsset{compat=1.16}

\usepackage{thmtools}
\usepackage{thm-restate}
\usepackage{mathtools}

\newtheoremstyle{myStyle}
{11pt} 
{11pt} 
{\itshape} 
{} 
{\bfseries} 
{.} 
{.5em} 
{} 

\theoremstyle{myStyle}

\definecolor{darkred}{rgb}{0.55, 0.0, 0.0}
\definecolor{darkorange}{rgb}{.7509, .2457, .0035}